\documentclass[runningheads,a4paper]{llncs}
\usepackage{amssymb}
\usepackage[pdftex]{graphicx}
\usepackage{amssymb}
\usepackage[utf8]{inputenc}
\usepackage{url}
\usepackage{float}
\usepackage{amsmath}
\usepackage{graphicx}
\usepackage{wrapfig}
\usepackage{booktabs}
\usepackage{multirow}
\usepackage{url}
\usepackage{lipsum}

\makeatletter
\def\Hline{%
\noalign{\ifnum0=`}\fi\hrule \@height 1pt \futurelet
\reserved@a\@xhline}
\makeatother

\begin{document}

\title{Hibikino-Musashi@Home\\2019 Team Description Paper}

\author{
Yuichiro Tanaka
\and Yutaro Ishida
\and Yushi Abe
\and Tomohiro Ono
\and Kohei Kabashima
\and Takuma Sakata
\and Masashi Fukuyado
\and Fuyuki Muto
\and Takumi Yoshii
\and Kazuki Kanamaru
\and Daichi Kamimura
\and Kentaro Nakamura
\and Yuta Nishimura
\and Takashi Morie
\and Hakaru Tamukoh
}
\institute{
Graduate school of life science and systems engineering,\\
Kyushu Institute of Technology,\\
2-4 Hibikino, Wakamatsu-ku, Kitakyushu 808-0196, Japan,\\
\texttt{http://www.brain.kyutech.ac.jp/\~{}hma/wordpress/}
}
\authorrunning{Yuichiro Tanaka et al.}
\maketitle

%%%%%%%%%%%%%%%%%%%%%%%%%%%%%%%%%%%%%%%%%%%%%%%%%%%%%%%%%%%%%%%%%%%%%%%%%%%%%%%%%%%%

\begin{abstract}
Our team, Hibikino-Musashi@Home (HMA), was founded in 2010.
It is based in the Kitakyushu Science and Research Park, Japan.
Since 2010, we have participated in the RoboCup@Home Japan Open competition open platform league annually.
We have also participated in the RoboCup 2017 Nagoya as an open platform league and domestic standard platform league teams, and in the RoboCup 2018 Montreal as a domestic standard platform league team.
Currently, we have 23 members from seven different laboratories based in Kyushu Institute of Technology.
This paper aims to introduce the activities that are performed by our team and the technologies that we use.
\end{abstract}

%%%%%%%%%%%%%%%%%%%%%%%%%%%%%%%%%%%%%%%%%%%%%%%%%%%%%%%%%%%%%%%%%%%%%%%%%%%%%%%%%%%%

\section{Introduction}
Our team, Hibikino-Musashi@Home (HMA) was founded in 2010 and it has been competing in the RoboCup@Home Japan Open competition open platform league (OPL) annually since then.
Our team is developing a home-service robot, and we intend to demonstrate our robot in this event to present our research outcomes.\par

In RoboCup 2017 Nagoya, we participated in OPL and domestic standard platform league (DSPL), and in the RoboCup 2018 Montreal, we participated in DSPL.
In both these competitions, we employed a TOYOTA HSR robot and were awarded the first prize in DSPL\cite{toyota_hsr}.

This paper describes the technologies used by us.
Especially, this paper outlines an object recognition system that uses deep learning \cite{hinton2006fast}, speech recognition system, sound localization system, and a brain-inspired amygdala model, which was originally proposed by us and is installed in our HSR.

%%%%%%%%%%%%%%%%%%%%%%%%%%%%%%%%%%%%%%%%%%%%%%%%%%%%%%%%%%%%%%%%%%%%%%%%%%%%%%%%%%%%

\section{System overview}
Figure \ref{fig:softOverview} presents an overview of our HSR system.
We has used an HSR since 2016.
In this section, we will introduce the specifications of our HSR.

\subsection{Hardware overview}
%HMA uses the HSR shown in Fig. \ref{fig:cust_hsr}.
We participated in RoboCup 2018 Montreal with the HSR.
The computational resources built into the HSR were inadequate to support our intelligent systems and were unable to extract the maximum performance from the system.
To overcome this limitation, using an Official Standard Laptop for DSPL that can fulfill the computational requirements of our intelligent systems has been permitted since RoboCup 2018 Montreal.
One of our team members (Yutaro ISHIDA) contributed a rule definition of the Official Standard Laptop for DSPL discussed in GitHub \cite{github366,github368}.
We use an ALIENWARE (Intel Core i7-8700K CPU, 32GB RAM and GTX-1080 GPU) as the Official Standard Laptop for DSPL.
Consequently, the computer equipped inside the HSR could be used to run basic HSR software, such as its sensor drivers and motion planning and actuator drivers.
This increased the operational stability of the HSR.
%
%\begin{figure}[tb]
%\begin{center}
%\includegraphics[scale=0.5]{./images/customizedHSR.png}
%\caption{Devices installed on HSR.}
%\label{fig:cust_hsr}
%\end{center}
%\end{figure}

\subsection{Software overview}
\begin{figure}[bt]
\begin{center}
\includegraphics[scale=0.5]{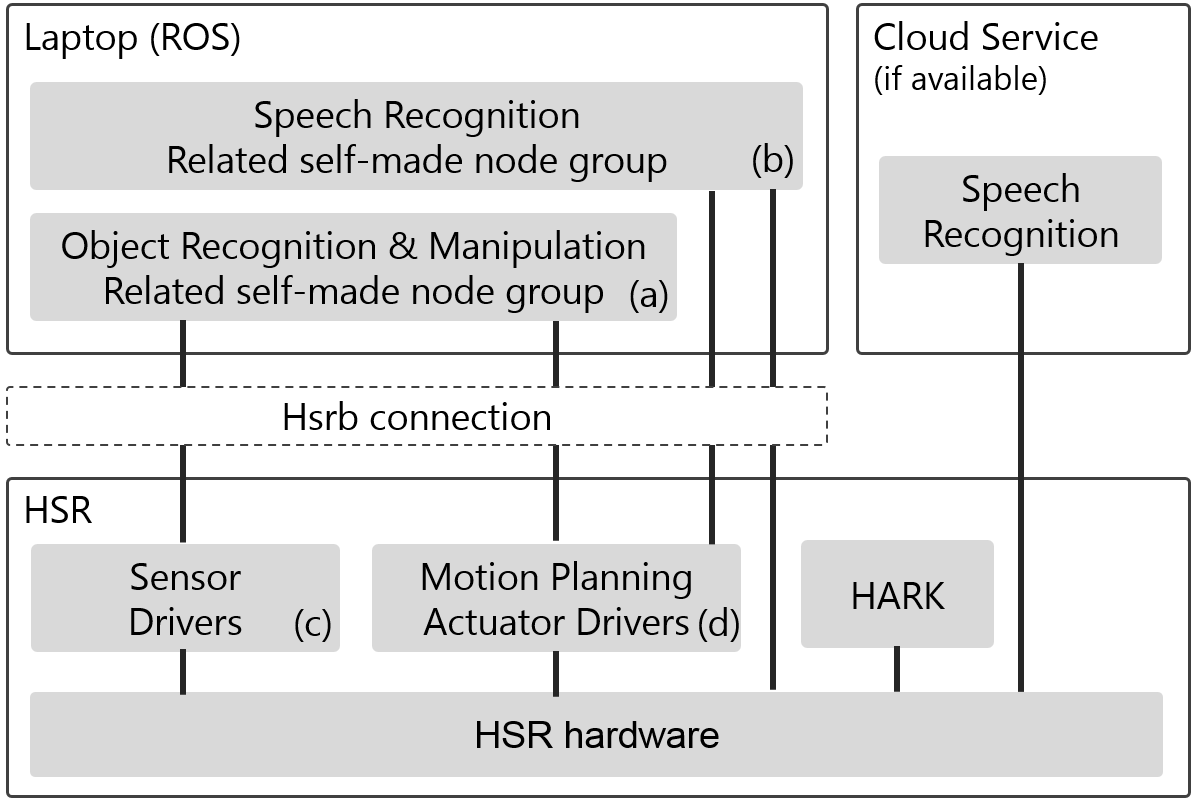}
\caption{Block diagram overview of our HSR system.}
\label{fig:softOverview}
\end{center}
\end{figure}
In this section, we introduce the software installed in our HSR.
Figure \ref{fig:softOverview} shows the system installed in our HSR.
The system is based on the Robot Operating System \cite{ros}.
In the our HSR system, the laptop computer and a cloud service, only if a network connection is available, are used for system processing.
The laptop is connected to a computer through an Hsrb interface.
The built-in computer specializes in low-layer systems, such as the HSR's sensor drivers, motion planning, and actuator drivers, as shown in Fig. \ref{fig:softOverview} (c)(d).

%%%%%%%%%%%%%%%%%%%%%%%%%%%%%%%%%%%%%%%%%%%%%%%%%%%%%%%%%%%%%%%%%%%%%%%%%%%%%%%%%%%%

\section{Object recognition}
In this section, we explain the object recognition system (shown in Fig, \ref{fig:softOverview} (a)), which is based on you look only once (YOLO) \cite{redmon2016you}.

To train YOLO, a complex annotation phase is required for annotating labels and bounding boxes of objects.
In the RoboCup@Home competition, predefined objects are usually announced during the setup days right before the start of the competition days.
Thus, we have limited time to train YOLO in the competition, and the annotation phase impedes the use of the trained YOLO in the competition days.

\begin{figure}[b]
\begin{center}
\includegraphics[scale=0.5]{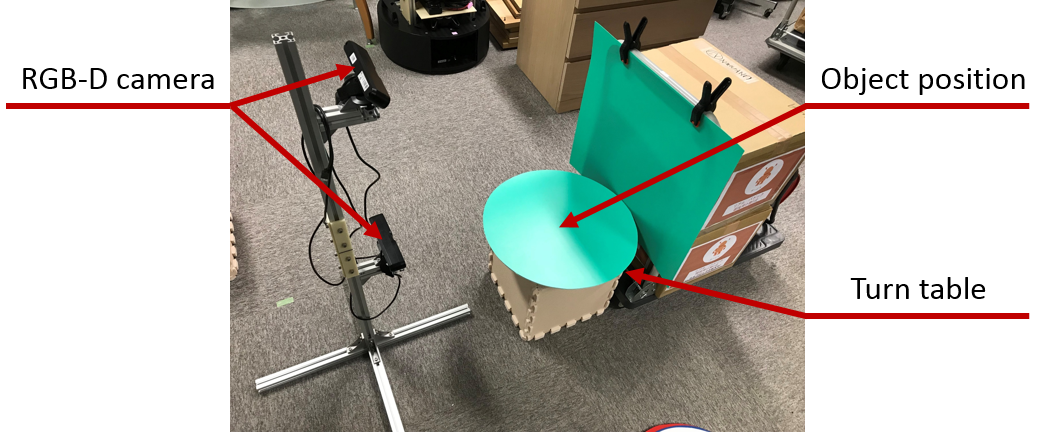}
\caption{Overview of proposed autonomous annotation system for YOLO.}
\label{fig:annotation1}
\end{center}
\end{figure}

We propose an autonomous annotation system for YOLO using chroma keys.
Figure \ref{fig:annotation1} shows an overview of the proposed system.
In the system, two RGB-D cameras are used, and they are identical to the camera implemented on the HSR.
These cameras differ in terms of mounting positions and angles: one of the cameras captures an object from a higher position than the other camera.
Since two different mounting positions and angles are used, two different images of a given object are captured simultaneously.
Using these images for training, YOLO can recognize objects captured from high and low positions.
The objects captured by the cameras are placed on a turntable.
We capture 200 images of a given object by one camera as the table turns.
Thus, we obtain 400 images per object.

\begin{figure}[tb]
\begin{center}
\includegraphics[scale=0.5]{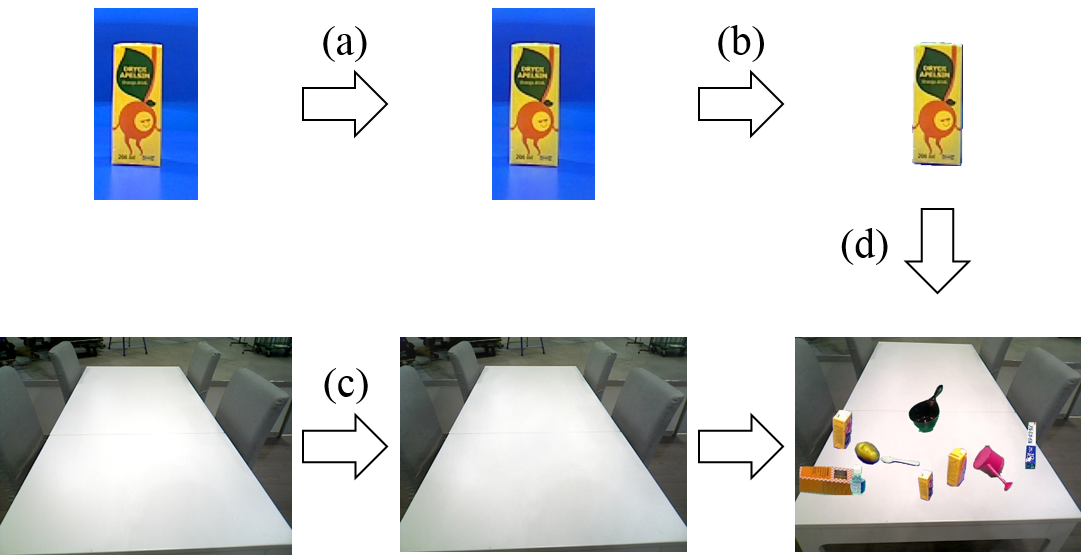}
\caption{Processing flow to generate training images for YOLO.}
\label{fig:annotation2}
\end{center}
\end{figure}

Figure \ref{fig:annotation2} shows the processing flow to generate training images for YOLO.
To adapt to various lightning conditions, we apply an automatic color equation algorithm \cite{RIZZI20031663} to the captured images (Fig. \ref{fig:annotation2} (a)).
We use a python library \cite{colorcorrect} to this end.
Then, we remove image backgrounds using chroma keys (Fig. \ref{fig:annotation2} (b)).

For backgrounds of training images, we shoot background images, for example, table and shelf, among others.
Moreover, we apply the automatic color equation algorithm to the background images (Fig. \ref{fig:annotation2} (c)).
To incorporate the object images into the background images, we define 20-25 object locations on the background images (the number of object locations depends on the background images).
Then, by placing the object images on the defined object locations autonomously, training images for YOLO are generated (Fig. \ref{fig:annotation2} (d)).
If there are 91 class objects and 100 background images, 1,237,600 training images are generated.
Additionally, annotation data for the training images are generated autonomously because object labels and positions are known.

Image generation requires ~30 min (in parallel using six CPU cores), and training of YOLO requires approximately six hours when using the GTX1080 GPU on the Standard Laptop.
Even though the generated training data are artificial, recognition of YOLO in actual environments works.

%%%%%%%%%%%%%%%%%%%%%%%%%%%%%%%%%%%%%%%%%%%%%%%%%%%%%%%%%%%%%%%%%%%%%%%%%%%%%%%%%%%%

\section{Speech recognition and sound localization}
The voice recognition/sound source localization system (shown in Fig. \ref{fig:softOverview} (b)) operates as follows:

\begin{enumerate}
\item The microphone array captures the voice of the person addressing it and uses a speech recognition engine, Web Speech API on google chrome, to recognize what is being said.
\item The voice of the person addressing the microphone array is captured, and sound source localization is performed according to the MUSIC method using HARK \cite{hark}, an auditory software package for robots.
\end{enumerate}

%%%%%%%%%%%%%%%%%%%%%%%%%%%%%%%%%%%%%%%%%%%%%%%%%%%%%%%%%%%%%%%%%%%%%%%%%%%%%%%%%%%%

\section{Brain-inspired amygdala model}
In this section, we explain the brain-inspired amygdala model that learns preferences through human-robot interactions.

Two types of knowledge are required by home-service robots: the first is common knowledge pertaining to the world and the second is local knowledge depending on environment.
For example, common knowledge is required when a robot is asked to bring "green tea."'
In this case, the robot must know what "green tea" is.
To obtain common knowledge, deep learning is one of the powerful solutions because big data on "green tea" is available.
On the contrary, local knowledge is required when a robot is asked to bring "that."
In this case, the robot must know what "that" is, and "that" depends on people's preferences.
To obtain local knowledge, deep learning is not effective because big data on "that" is unavailable.
In the case of humans, we can know someone's preference from our past experiences with that human.
Similarly, the robot must obtain such preferences from a few human-robot interactions.

We focus on the amygdala, an area of the human brain.
The amygdala causes fear conditioning \cite{Ledoux2003}, a type of classical conditioning that has been made popular by Pavlov's dogs.
By applying classical conditioning to home-service robots, the robots can be made to obtain local knowledge through a few human-robot interactions, similar to that in Pavlov's experiments with dogs.

\begin{figure}[bt]
\begin{center}
\includegraphics[scale=0.5]{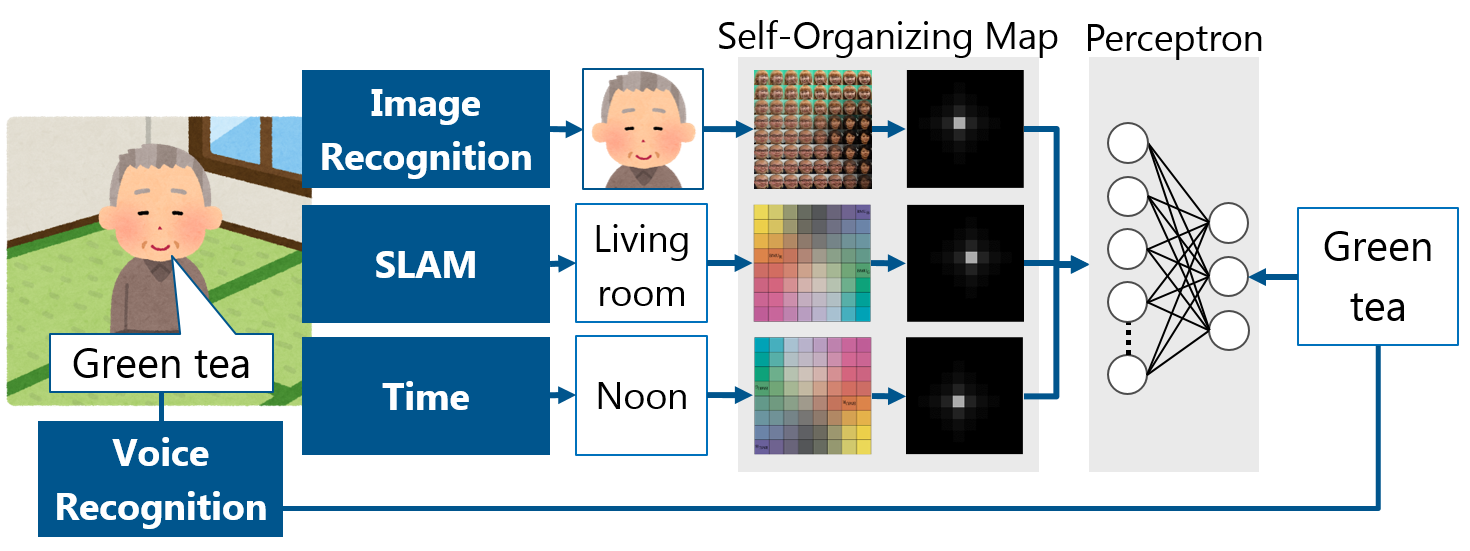}
\caption{Brain-inspired amygdala model.}
\label{fig:amygdala1}
\end{center}
\end{figure}

Figure \ref{fig:amygdala1} shows the proposed amygdala model \cite{tanaka2018amygdala}.
The model comprises multiple self-organizing maps (SOMs) \cite{Kohonen1982} and a single perceptron.
When a robot is asked to bring an object, the model receives information about the ordered object via voice recognition.
At the same time, the model receives information about the face of the person placing the order via image recognition and about their location via SLAM and time.
These pieces of information are input into the SOMs.
Then, the SOMs classify the information and output the classification results.
The classification results are input into the perceptron, and the perceptron learns relation between the classification results and the ordered object.
Thereafter, if the robot is asked to bring "that," the model can estimate what the ordering person wants by using information about face, place, and time.

We confirmed experimentally that the model learns preferences through a few human-robot interactions.
In the experiment, we defined two situations; A and B.
In situation A, face A, place A, and time A are always given, and the ordered object is always object A.
In situation B, face A, place B, and time B are always given, and ordered object is always object B.
At first, we input situation A into the model as an interaction and repeated the same procedure five times.
Then, we input situation B into the model as an interaction and repeated the same procedure five times.

\begin{figure}[tb]
\begin{center}
\includegraphics[scale=0.65]{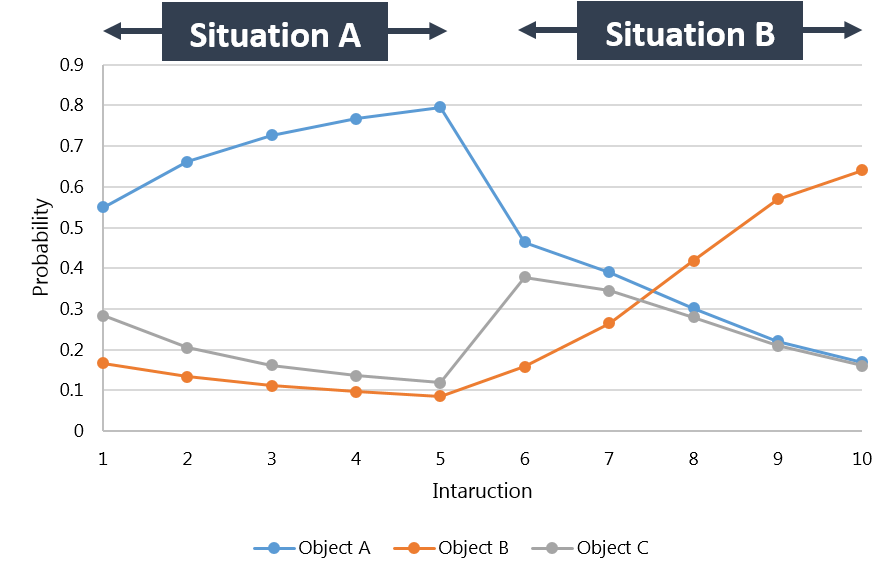}
\caption{Learning preferences through human-robot interactions.}
\label{fig:amygdala2}
\end{center}
\end{figure}

Figure \ref{fig:amygdala2} shows the results of our experiments.
The vertical axis in the figure indicates the probability that the model estimates preferences from face, place, and time.
During the interaction associated with situation A, the probability of object A increases.
On the contrary, during the interaction associated with situation B, the probability of object B become increases and peaks in the eighth interaction.
Thus, the model learns preferences from a few human-robot interactions.

%%%%%%%%%%%%%%%%%%%%%%%%%%%%%%%%%%%%%%%%%%%%%%%%%%%%%%%%%%%%%%%%%%%%%%%%%%%%%%%%%%%%

\section{Competition results}

\begin{table}[t]
\begin{center}
\caption{Results of recent competitions.}
\label{tab:result}
\begin{tabular}{l|l} \Hline
	\multicolumn{1}{c|}{Competition}	& \multicolumn{1}{c}{Result}			\\ \Hline
	RoboCup Japan Open 2017 Aichi		& @Home DSPL 2nd				\\
        					& @Home OPL 3rd					\\ \hline
	RoboCup 2017 Nagoya			& {\bf @Home DSPL 1st}				\\
        					& @Home OPL 5th					\\ \hline
        RoboCup Japan Open 2018 Ogaki		& @Home DSPL 2nd				\\
        					& @Home OPL 1st					\\
                                        	& JSAI Award					\\ \hline
        RoboCup 2018 Montreal			& {\bf @Home DSPL 1st}				\\
        					& P\&G Dishwasher Challenge Award		\\ \hline
        World Robot Challenge 2018		& Service Robotics Category 			\\
        					& Partner Robot Challenge Real Space 1st 	\\
                                        	& METI Minister's Award 			\\
                                        	& RSJ Special Award				\\ \Hline
\end{tabular}
\end{center}
\end{table}

Table \ref{tab:result} shows the results achieved by our team in the recent competitions.
We participated in RoboCup and World Robot Challenge for several years, and as a result, our team has won prizes and academic awards. \par
Especially, we participated in the RoboCup 2018 Montreal using the system described herein.
We scored 335 points out of 2125 points.
These points were 62 \% of the points scored by the top OPL team.
We were able to demonstrate the performance of HSR and our technologies.
Especially, we won the Procter \& Gamble Dishwasher Challenge Award in RoboCup 2018 owing to the object recognition and manipulation used by YOLO.
Thanks to these results, we were awarded the first prize in the competition.

%%%%%%%%%%%%%%%%%%%%%%%%%%%%%%%%%%%%%%%%%%%%%%%%%%%%%%%%%%%%%%%%%%%%%%%%%%%%%%%%%%%%

\vspace{-0.3cm}
\section{Conclusions}
In this paper, we summarized available information about our HSR, which we entered into RoboCup 2018 Montreal.
The object recognition and voice interaction capabilities that we built into the robot were described as well.
Currently, we are developing many different pieces of software for an HSR that will be entered into RoboCup 2019 Sydney.

\vspace{-0.3cm}
\section*{GitHub}
The source codes of our systems and our original dataset are published on GitHub. The URL is as follows:\\
https://github.com/hibikino-musashi-athome

\vspace{-0.3cm}
\section*{Acknowledgment}
This work was supported by Ministry of Education, Culture, Sports, Science and Technology, Joint Graduate School Intelligent Car \& Robotics course (2012-2017),
Kitakyushu Foundation for the Advancement of Industry Science and Technology (2013-2015),
Kyushu Institute of Technology 100th anniversary commemoration project : student project (2015, 2018) and YASKAWA electric corporation project (2016-2017),
JSPS KAKENHI grant number 17H01798,
and the New Energy and Industrial Technology Development Organization (NEDO).

\newpage
%\vspace{-0.3cm}
\bibliography{tdp2019}
\bibliographystyle{unsrt}

\newpage
%\section*{Robot HSR Hardware Description}
%\textit{(In this section briefly describe the software and hardware of the robot)}
%In this section briefly describe the hardware of the robot

%\begin{itemize}
%	\item Name: Human support robot (HSR).
%	\item Footprint: 430 mm.
%	\item Height(min/max): 1005/1350 (top of the head height).
%	\item Weight: About 37 kg.
%	\item Sensors on the moving base:
%		\begin{itemize}
%			\item Laser range sensor.
%			\item IMU.
%		\end{itemize}
%	\item Arm length: 600 mm.
%	\item Arm payload (recommended/max): 0.5/1.2 kg.
%	\item Sensors on the gripper:
%		\begin{itemize}
%			\item Gripping force sensor.
%			\item Wide-angle camera.
%		\end{itemize}
%	\item Sensors on the head:
%		\begin{itemize}
%			\item RGB-D sensor.
%			\item Stereo camera.
%			\item Wide-angle camera.
%			\item Microphone array.
%		\end{itemize}
%	\item Body expandability: USB x3, VGA x1, LAN x1, Serial x1, and 15V-0.5A output x1.
%\end{itemize}

%\textit{Also our robot incorporates the following devices:}

\section*{Robot's Software Description}
For our robot we are using the following software:

\begin{itemize}
	\item OS: Ubuntu 16.04.
	\item Middleware: ROS Kinetic.
	\item State management: SMACH (ROS).
	\item Speech recognition (English):
		\begin{itemize}
			\item rospeex \cite{rospeex}.
			\item Web Speech API.
			\item Julius.
		\end{itemize}
	\item Morphological Analysis Dependency Structure Analysis (English): SyntaxNet.
	\item Speech synthesis (English): Web Speech API.
	\item Speech recognition (Japanese): Julius.
	\item Morphological Analysis (Japanese): MeCab.
	\item Dependency structure analysis (Japanese): CaboCha.
	\item Speech synthesis (Japanese): Open JTalk.
	\item Sound location: HARK.
	\item Object detection: Point cloud library (PCL) and You only look once (YOLO) \cite{redmon2016you}.
	\item Object recognition: YOLO.
	\item Human detection / tracking:
		\begin{itemize}
			\item Depth image + particle filter.
			\item OpenPose \cite{cao2017realtime}.
		\end{itemize}
	\item Face detection: Convolutional Neural Network.
	\item SLAM: hector\_slam (ROS).
	\item Path planning: move\_base (ROS).
\end{itemize}

\end{document}